\pdfoutput=1

\documentclass[11pt]{article}

\usepackage[]{acl}

  \usepackage{times}
  \usepackage{latexsym}
  \usepackage{multirow}
  \usepackage{multicol}
  \usepackage{amsmath}
  \usepackage{graphicx}
  \usepackage{enumitem}
  \usepackage{adjustbox}
  \usepackage{makecell}
  \usepackage{listings}
  \usepackage[T1]{fontenc}

  \usepackage[utf8]{inputenc}

  \usepackage{microtype}

  \usepackage{inconsolata}

\title{Test-Time Code-Switching for Cross-lingual Aspect Sentiment Triplet Extraction}

\author{Dongming Sheng \quad Kexin Han \quad Hao Li \quad Yan Zhang\\ \quad \bf Yucheng Huang \quad \bf Jun Lang \quad \bf Wenqiang Liu \\
        Tencent \quad \\
        \texttt{\{hanssheng,masonqliu\}@tencent.com} \\
}

\begin{document}
\maketitle
\begin{abstract}
  Aspect Sentiment Triplet Extraction (ASTE) is a thriving research area with impressive outcomes being achieved on high-resource languages. However, the application of cross-lingual transfer to the ASTE task has been relatively unexplored, and current code-switching methods still suffer from term boundary detection issues and out-of-dictionary problems. In this study, we introduce a novel \textbf{T}est-\textbf{T}ime \textbf{C}ode-\textbf{SW}itching (\textbf{TT-CSW}) framework, which bridges the gap between the bilingual training phase and the monolingual test-time prediction. During training, a generative model is developed based on bilingual code-switched training data and can produce bilingual ASTE triplets for bilingual inputs. In the testing stage, we employ an alignment-based code-switching technique for test-time augmentation. Extensive experiments on cross-lingual ASTE datasets validate the effectiveness of our proposed method. We achieve an average improvement of 3.7\% in terms of weighted-averaged F1 in four datasets with different languages. Additionally, we set a benchmark using ChatGPT and GPT-4, and demonstrate that even smaller generative models fine-tuned with our proposed TT-CSW framework surpass ChatGPT and GPT-4 by 14.2\% and 5.0\% respectively.
\end{abstract}

\section{Introduction}
  Aspect sentiment Triplet Extraction (ASTE) task has drawn increasing attention in recent years \citep{peng2020knowing, xu2020position, zhang2023target, li2023dual}. It aims at the co-extraction of aspect terms, opinion terms and sentiment polarities. Despite the success achieved on high-resource languages, it is still challenging to attain comparable performance for languages with limited annotation resources. This highlights the need for cross-lingual ASTE, an extended task commonly trained on languages with rich annotation resources (e.g., English) and tested on those with low resources (e.g., Basque and Catalan). 
    
  \begin{figure}[!t]
    \centering \includegraphics[width=0.99\linewidth]{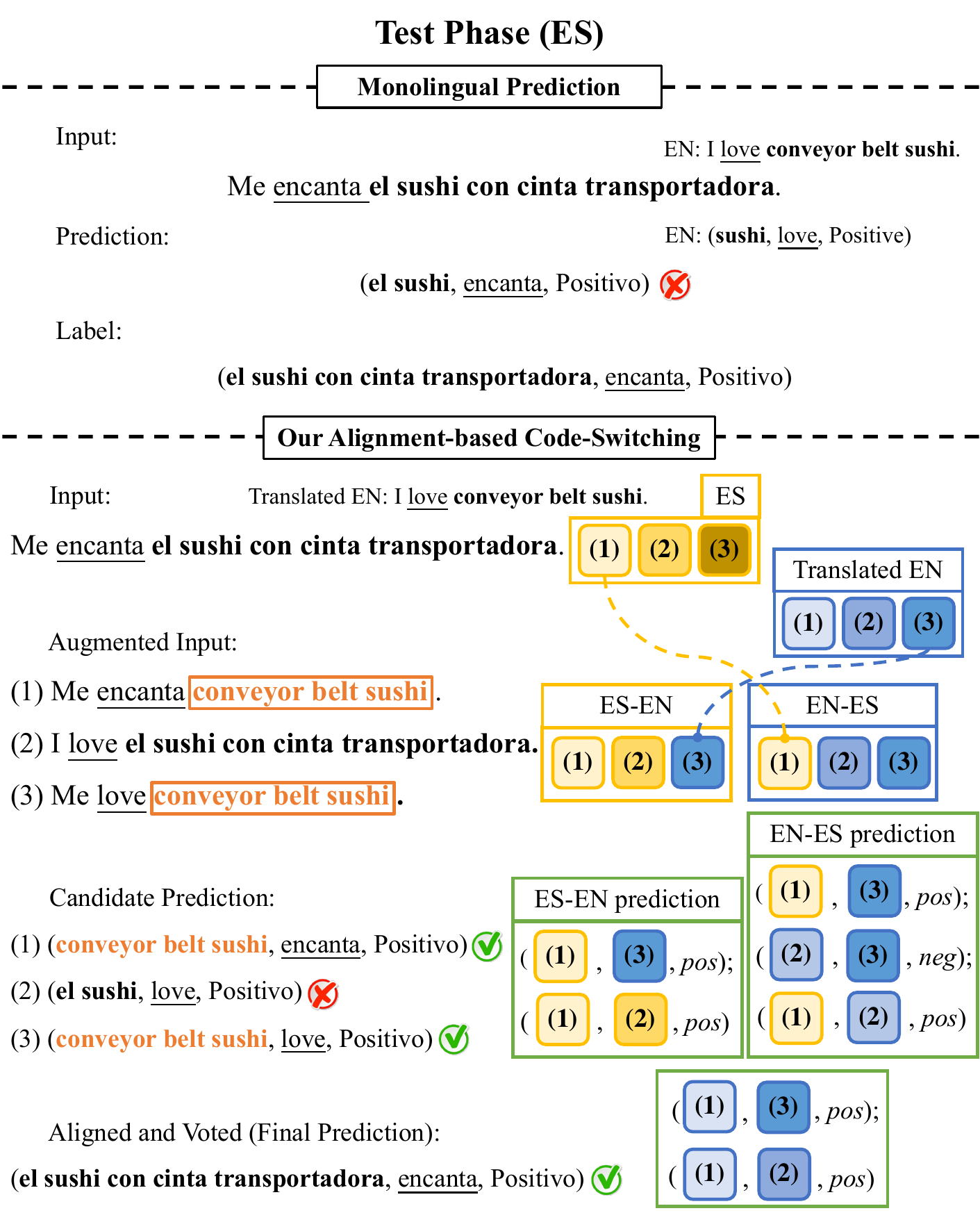}
    \caption{An example of testing phase in cross-lingual ASTE task on Spanish dataset. Phrases with bold and underlined words represent \textbf{aspect} and \underline{opinion} terms respectively. The substituted words are highlighted within the orange boxes. The diagram on the bottom right illustrates the pipeline of our proposed alignment-based code-switching method.}
    \label{fig:intro_example}
  \end{figure}

  Recent studies have demonstrated that code-switching can effectively facilitate cross-lingual transfer for low-resource languages across various NLP tasks \citep{li2022multi, zhang2021cross, qin2021cosda, zhu2023enhancing}. However, in terms of cross-lingual ASTE task, current code-switching methods still suffer from two major issues.


  Firstly, existing code-switching techniques are mainly used during the training phase as a method of data augmentation \citep{li2022multi, zhang2021cross}, and the prediction is only based on the target language. However, the detection of term boundaries tends to be a challenge when conducting inference in a monolingual context for languages with limited annotation resources. As shown in Figure \ref{fig:intro_example}, \textit{el sushi con cinta transportadora} means \textit{conveyor belt sushi}(i.e., a type of sushi restaurant) in English. Due to limited annotation resources, the model may fail to recognize the aspect term \textit{el sushi con cinta transportadora} as a whole, and instead predict \textit{el sushi} only. This leads to the incorrect prediction of term boundaries.


  Furthermore, in the context of the ASTE task, both opinion and aspect terms typically form phrases. This characteristic poses a challenge for existing code-switching methods that rely on bilingual dictionaries \citep{qin2021cosda, feng2022toward} or follow the translate-then-align procedure \citep{mayhew2017cheap, fei2020cross}. For instance, \textit{sound insulation} is not present in the bilingual dictionary, and translating \textit{insulation} directly could result in semantic inaccuracies. Moreover, these terms can frequently be proper nouns, like the English brand name \textit{Hard Rock Cafe}. However, these terms often fall outside the scope of bilingual dictionaries (i.e., out-of-dictionary issues), leading to issues of inconsistency and inaccuracies due to incorrect translations.


  Based on the above observations, we propose a \textbf{T}est-\textbf{T}ime \textbf{C}ode-\textbf{SW}itching framework (\textbf{TT-CSW}) for the cross-lingual ASTE task. In our framework, the code-switching method offers a bilingual context in both training and testing phases. In this way, our framework can act as a bridge between the monolingual test-time prediction and the bilingual training phase.

  For the training stage, our model learns to predict bilingual ASTE triplets based on code-switched inputs. To deal with the issue of out-of-dictionary and term boundary detection, we propose a boundary-aware code-switching method. This approach preserves the completeness of aspect and opinion terms during the translation process, circumventing problems associated with inconsistency and inaccuracies due to wrong translations from bilingual dictionaries. Consequently, it considerably enhances the alignment capability for models to understand bilingual context and predict term boundaries accurately.

  During testing stage, our model can utilize knowledge from code-switching to generate triplets in the target language. To further address the incorrect prediction for term boundaries during test-time, we introduce a code-switching method based on alignment for test-time augmentation, as illustrated in Figure \ref{fig:intro_example}. A heuristic switching strategy is designed to generate a set of code-switched augmentation examples. The output triplets from these examples are then aligned into the target language for the final output. The integration of code-switching during the testing stage provides a bilingual multi-view of the input sentence, which incorporates information from source language with rich annotation resources, and improves performance for predicting term boundaries.

  Extensive experiments on cross-lingual ASTE datasets validate the effectiveness of our proposed method. By integrating our method with various backbone models, we achieve an average improvement of 3.7\% in terms of weighted-averaged F1 in four datasets with different languages. Furthermore, we benchmark ChatGPT \footnote{https://chat.openai.com/} and GPT-4 \footnote{https://openai.com/gpt-4}, OpenAI's widely-used Large Language Model (LLM) and illustrate that small generative models finetuned with our proposed TT-CSW framework still outperform ChatGPT by 14.2\% and 5.0\% respectively in terms of weighted-averaged F1.
  
  Our main contributions are as follows: 
  
  1) We propose a novel test-time code-switching framework for cross-lingual ASTE task, which can be easily integrated with various backbone models. 
  
  2) We develop a boundary-aware code-switching method based on translation system for solving the issue of out-of-dictionary and phrase code-switching. 
  
  3) We design an alignment method for test-time augmentation to improve term boundary prediction. 
  
  4) We benchmark ChatGPT and GPT-4 on cross-lingual ASTE task and show that small generative models finetuned with our TT-CSW framework can still outperform ChatGPT and GPT-4.

\section{Methodology}


  \begin{figure*}[ht]
    \centering \includegraphics[width=0.99\linewidth]{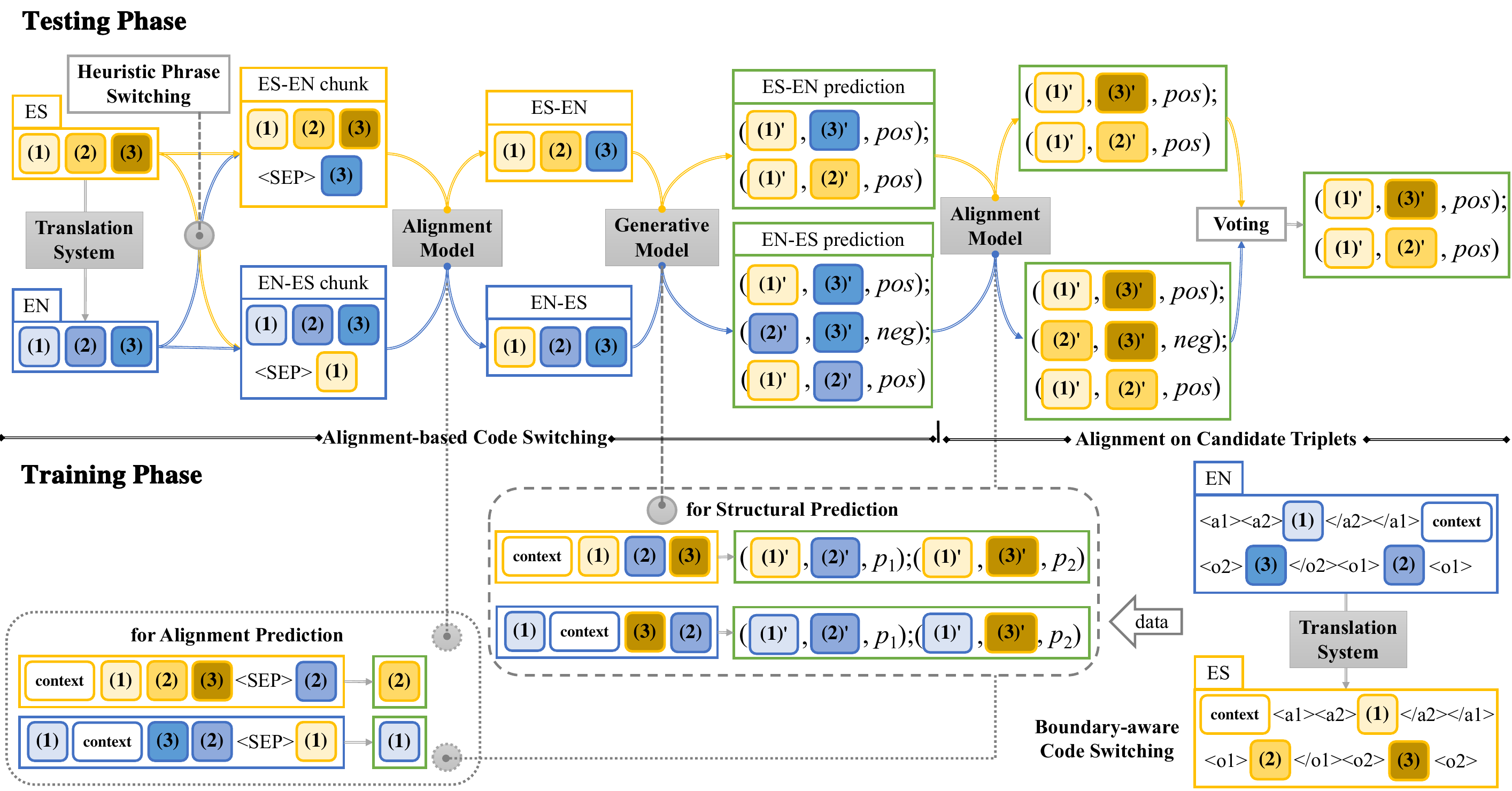}
    \caption{The overall architecture of our proposed TT-CSW framework.}
    \label{fig:model_arch}
  \end{figure*}

  Our proposed TT-CSW framework, as depicted in Figure \ref{fig:model_arch}, is made up of two key components: the training phase and the testing phase. The structure of the training phase, shown on the bottom side, is composed of three elements: boundary-aware code-switching, structural prediction, and alignment prediction. The layout of the testing phase, as illustrated on the upper side, involves two parts: alignment-based code-switching, and alignment of candidate triplets. In this section, we begin by providing a formal definition of the cross-lingual ASTE task, then we delve into the specifics of our proposed TT-CSW framework.


  \subsection{Task Definition}
    \label{sec:task_def}
    We denote an monolingual ASTE dataset with $N$ samples as $D = \{D_1, D_2, ..., D_N\}$. For each sample $D_i = \{s_i, RT_i\}$, $s_i$ represents the input sentence, and $RT_i=\{T_1, ..., T_n\}$ represents the ground truth triplet list for the input sentence. Each triplet $T_i=\{a_i, o_i, p_i\}$ consists of aspect term ($a$), opinion term ($o$) and sentiment polarity ($p$). The ASTE task aims to predict a list of $m$ predicted triplets $PT_i=\{T_1, T_2, ..., T_m\}$ for each of the input sentence $s_i$. For cross-lingual ASTE, we denote the dataset for source language as $D^{\left(src\right)}$, and the dataset for target language as $D^{\left(tgt\right)}$. We need to train our model $M$ on $D^{\left(src\right)}$ and perform inference on $D^{\left(tgt\right)}$.

  \subsection{Training Phase}
    For the training phase, we first create a bilingual code-switching dataset using our proposed boundary-aware code-switching method. Following this, we train two separate models utilizing this dataset: a generative bilingual model and a bilingual alignment model. The generative bilingual model is designed to produce bilingual ASTE triplets grounded on the bilingual context. Meanwhile, the role of bilingual alignment model is to convert the bilingual candidate triplets into the same language during the testing phase.

    \subsubsection{Boundary-aware Code-switching}
      Traditional methods for creating code-switching context rely on bilingual dictionary \citep{qin2021cosda, feng2022toward}, or follow the procedure of translate-then-align \citep{mayhew2017cheap, fei2020cross}, which utilizes word alignment tools after translation. Inspired by \citet{zhang2021cross}, we propose a boundary-aware code-switching method via the translation system without the use of bilingual dictionaries or word alignment tools, as shown on the bottom right of Figure \ref{fig:model_arch}. 

      Given that there are often multiple triplets within a single sentence, we introduce the HTML tags to locate these triplets. We employ <$a_i$> and <$/a_i$> to indicate the start and end of the $i$-th aspect term, and <$o_i$> and <$/o_i$> to denote the boundaries of the $i$-th opinion term. Therefore, we can distinctively differentiate multiple aspect and opinion term pairs during the translation, while preserving their original boundaries. With the help of the HTML tags, we can easily construct bilingual parallel phrases for training the bilingual alignment model in section \ref{sec:alignment_model}. Since the HTML tags are not part of the original sentence, we can easily remove them after the translation. The sentiment polarity remains unchanged during the translation, providing extra training examples for both the bilingual generative model and the bilingual alignment model.

    \subsubsection{Bilingual Structural Prediction}
      We develop a bilingual generative model trained on the boundary-aware code-switching dataset to perform the structural prediction task, as shown on the bottom center of Figure \ref{fig:model_arch}. To serialize the output triplets, we opt not to use commas or semicolons as separators or connectors within and between triplets, which is a standard practice in the GAS-Extraction format \citep{zhang2021towards}. The reason for this is that these symbols could also be present in aspect or opinion terms, leading to format confusion. Instead, we choose to use the special tokens <split> and <join> for this purpose. As an example, for a input sequence with two triplets ($a_1$, $o_1$, $p_1$) and ($a_2$, $o_2$, $p_2$), we use the following format for structural prediction: ($a_1$<split>$o_1$<split>$p_1$) <join> ($a_2$<split>$o_2$<split>$p_2$).

    \subsubsection{Bilingual Alignment Prediction}
      \label{sec:alignment_model}
      We use bilingual parallel phrases from boundary-aware code-switching dataset to train a bilingual alignment model, which will be used for testing phase in section \ref{sec:testing_phase}. We use the mT5-base model as the backbone. For a pair of parallel aspect term $a$ and $a^{\left(T\right)}$, we append the translated term $a^{\left(T\right)}$ to the end of original sentence and use a special token <SEP> to separate them, as depicted on the bottom left of Figure \ref{fig:model_arch}. We enhance computational efficiency and improve diversity by segmenting the original sentence into multiple chunks, each with a maximum length of 128, using a sliding window method. If the sentence chunk does not contain the original term $a$ after segmentation, we treat it as a negative sample by setting the ground-truth label to \textit{None}. Additionally, to further generate negative samples and improve robustness of our alignment model, we substitute 10\% of the translated term $a^{\left(T\right)}$ with a random token from the vocabulary. The ground-truth label for these randomly substituted terms is also set to \textit{None}.

  \subsection{Testing Phase}
    \label{sec:testing_phase}
    During the testing phase, we initially leverage the bilingual alignment model developed during the training phase to generate code-switch augmented examples. Subsequently, we utilize the bilingual generative model to produce a group of bilingual candidate triplets. Finally, using the bilingual alignment model, we align these candidate triplets into the target language. Through candidate voting, we obtain the final output triplets. For ease of demonstration, we denote the input sentence in target language during testing phase as $s^{\left(tgt\right)}$, and the input sentence in source language as $s^{\left(src\right)}$. 

    \subsubsection{Alignment-based Code-switching}
      \label{sec:alignment_csw}
      As depicted on the upper part of Figure \ref{fig:model_arch}, our code-switching method for test-time augmentation contains four steps. Initially, we use an off-the-shelf translation system to convert the target language into English. Then, we construct alignment inputs for the bilingual alignment model. A heuristic method is designed to select phrases with a maximum of 3-grams from the translated sentence. The criterion for selection is that bilingual alignment model should not predict \textit{None} as output. The top-10 longest phrases are chosen to construct the alignment inputs. There are two types of alignment inputs: $s^{\left(tgt\right)}$ <SEP> $t^{\left(src\right)}$ and $s^{\left(src\right)}$ <SEP> $t^{\left(tgt\right)}$, where $t^{\left(src\right)}$ and $t^{\left(tgt\right)}$ represent aspect term or opinion term in source language and target language respectively. Subsequently, we use the bilingual alignment model to get the augmented code-switching sentences. Finally, these sentences are used to create a set of candidate triplets with the help of the bilingual generative model.

    \subsubsection{Alignment on Candidate Triplets}
      As discussed in section \ref{sec:alignment_csw}, we generate two types of alignment inputs: $s^{\left(tgt\right)}$<SEP>$t^{\left(src\right)}$ and $s^{\left(src\right)}$<SEP>$t^{\left(tgt\right)}$. Consequently, we manage to create candidate triplets in two distinct languages. The former situation is straightforward because we aim for the triplets to be in the target language, and we\'ve already computed the bilingual parallel terms. However, for the latter situation, we still need to align the remaining terms from the source language into the target language. For this alignment, we continue to employ the bilingual alignment model that we obtain during the training time to ensure that the candidate triplets conform to a single language. Finally, to get the final output triplets, we use a voting mechanism to decide which of the terms in the candidate triplets are the most likely to be correct.

\section{Experiments} 
  \begin{table*}[!ht]
    \centering
    \begin{adjustbox}{width=0.97\textwidth}
    \begin{tabular}{|c|c|c|c|c|c|c|c|c|c|c|}
    \hline
        \multirow{2}{*}{Dataset} & \multirow{2}{*}{Language} & \multicolumn{3}{c|}{Train} & \multicolumn{3}{c|}{Validation} & \multicolumn{3}{c|}{Test} \\ \cline{3-11}
        & & \# $s$ & \# $a$ & \# $o$ & \# $s$ & \# $a$ & \# $o$ & \# $s$ & \# $a$ & \# $o$ \\ \hline
        \textbf{OpeNER}$_\textsuperscript{EN}$ & English & 2494 & 3850 & 4150 & 2494 & 3850 & 4150 & 2494 & 3850 & 4150\\ \hline      
        \textbf{NoReC}$_\textsuperscript{Fine}$ & Norwegian & - & - & - & - & - & - & 11437 & 8923 & 11115 \\ \hline
        \textbf{MultiB}$_\textsuperscript{EU}$ & Basque & - & - & - & - & - & - & 1521 & 1775 & 2328\\ \hline
        \textbf{MultiB}$_\textsuperscript{CA}$ & Catalan & - & - & - & - & - & - & 1678 & 2336 & 2756 \\ \hline
        \textbf{OpeNER}$_\textsuperscript{ES}$ & Spanish & - & - & - & - & - & - & 2057 & 3980 & 4388 \\ \hline
    \end{tabular}
    \end{adjustbox}
    \caption{\label{tab:dataset}
    Dataset statistics. \# $s$,  \# $a$ and \# $o$ refer to the number of sentences, the number of aspect terms, and the number of opinion terms respectively. For cross-lingual evaluation, training and validation sets are not available for the non-English datasets, which we denote as '-'.
    }
  \end{table*}

  \subsection{Datasets}
    We conduct experiments on the publicly available datasets from Semeval-2022 task 10: structured sentiment analysis \citep{barnes2022semeval}. We use the English OpeNER dataset \citep{agerri2013opener} for training, and perform cross-lingual validation on four datasets in other languages respectively, i.e., Spanish \citep{agerri2013opener}, Catalan \citep{barnes-etal-2018-multibooked}, Basque \citep{barnes-etal-2018-multibooked} and Norwegian \citep{ovrelid-etal-2020-fine} datasets. For reproducibility, we use the same train/validation/test split as the official datasets. The statistics of the datasets are listed in Table \ref{tab:dataset}. 

    The original datasets contain four types of annotations: holders, targets, expressions and polarities. As for the task of aspect sentiment triplet extraction, we treat the target annotations as aspect terms ($a$), and the expression annotations as opinion terms ($o$). The holder annotations are not used in our experiments. We did not use the multi-lingual dataset released in Semeval-2016 task 5 \citep{pontiki-etal-2016-semeval} because it does not contain opinion term annotations.

  \subsection{Implementation Details}
    All our experiments are conducted on a single NVIDIA Tesla P40 GPU with 24GB of GPU memory. We set the maximum sequence length to 128 and the training batch size to 8. We use AdamW optimizer with a learning rate of 1e-4. The model is trained for 10 epochs and checkpoints with the best performance on validation set are selected for the final predictions on test set. We use Google translate API \footnote{https://translate.google.com/} as the translation model in our experiments.

  \subsection{Compared Methods}
    We use the following models in our experiments: 
    \paragraph{mT5-base \citep{xue2021mt5}} mT5 is a multilingual variant of T5 \citep{raffel2020exploring}. T5 is a large-scale pre-trained language model with encoder-decoder architecture, and is trained with the span corruption task. mT5-base is pre-trained on a new Common Crawl-based dataset (mC4) covering 101 languages.
    
    \paragraph{m2m100\_418M \citep{fan2021beyond}} m2m100 is a variant of mBART \citep{liu2020multilingual}. mBART is a multi-lingual sequence-to-sequence model aimed for machine translation task. Compared to mBART, m2m100 is designed to be a many-to-many multilingual translation model that can translate directly between any pair of 100 languages. It is trained with the sequence-to-sequence denoising auto-encoding task.

    For performing cross-lingual ASTE task on m2m100, we need to append an additional language token to the input sentence, and set target language id to be the first generated token. As the original settings in m2m100 does not consider the bilingual code-switched context, we manually set the source language as English, and we use the target language id as the first generated token when generating triplets. We use the spanish language id as the first generated token for Basque dataset, for the reason that the m2m100 model does not contain Basque language id.
    
    \paragraph{ChatGPT \& GPT-4} ChatGPT and GPT-4 are large language models developed by OpenAI. ChatGPT is trained with both supervised fine-tuning (SFT) and Reinforcement Learning from Human Feedback (RLHF). GPT-4 surpasses ChatGPT in its advanced reasoning capabilities. It can solve difficult problems with greater accuracy, resulting from its broader general knowledge and problem solving abilities.
 
  \subsection{Evaluation Metrics}
    We use weighted-averaged precision, recall and F1-score to evaluate the performance of our model. Our evaluation metrics are calculated in the same way as Sentiment Graph F1 \citep{barnes2021structured}, with the exception that we do not utilize the graph-based structure for triplet representation. This graph-based structure requires additional alignment for generative models, which we sidestep by directly forming pairs based on the number of overlapping words between the predicted triplets and the ground-truth triplets. When calculating the precision score, we identify the most similar ground-truth triplet for each predicted triplet. Conversely, when calculating the recall score, we focus on pairing each ground-truth triplet with the most similar predicted triplet. The details of calculation can be found in \ref{sec:eval_metrics}.

  \subsection{Main Results}
    \begin{table*}[!ht]
      \centering
      \begin{adjustbox}{width=0.97\textwidth}
      \begin{tabular}{c|l||ccc|ccc|ccc|ccc||c}
      \hline
          \multicolumn{2}{c||}{} & \multicolumn{3}{c|}{Spanish} & \multicolumn{3}{c|}{Basque} & \multicolumn{3}{c|}{Catalan} & \multicolumn{3}{c||}{Norwegian} & \multirow{2}{*}{\textbf{AVG}} \\
          \multicolumn{2}{c||}{} & wP & wR & wF1 & wP & wR & wF1 & wP & wR & wF1& wP & wR & wF1 & \\ \hline

          \multicolumn{2}{c||}{all-null} & 11.7 & 4.8 & 6.8 & 21.3 & 12.9 & 16.1 & 16.1 & 9.4 & 11.8 & 47.0 & 32.6 & \textbf{38.5} & 18.3  \\ \hline

          \multirow{2}{*}{ChatGPT} & -0 & 42.7  & 35.8  & 38.9  & 25.2  & 19.1  & 21.8  & 39.6  & 34.7  & 37.0  & 25.3  & 26.5  & 25.9  & 30.9  \\
          & -10 & 48.9  & 48.2  & 48.5  & 24.0  & 26.8  & 25.3  & 41.3  & 41.9  & 41.6  & 12.4  & 15.7  & 13.9  & 32.3  \\ \hline

          GPT-4 & -10 & 61.0 & 50.3 & \textbf{55.1} & 33.4 & 29.2 & 31.2 & 48.8 & 43.3 & 45.9 & 38.3  & 30.7  & 34.1  & 41.6  \\ \hline

          \multirow{5}{*}{\makecell[c]{mT5-base\\ \citep{xue2021mt5}}} & CL & 45.5  & 29.4  & 35.7  & 35.1  & 28.1  & 31.2  & 42.6  & 34.5  & 38.1  & 39.3  & 33.6  & 36.2  & 35.3 \\ \cline{2-15}

          & CT & 52.9  & 42.9  & 47.4  & 44.1  & 42.6  & 43.3  & 49.8  & 47.7  & 48.7  & 32.7  & 31.2  & 31.9  & 42.8  \\ 
          & \multicolumn{1}{c||}{+tta} & 54.6  & 44.5  & 49.1  & 46.2  & 42.6  & \underline{44.4}  & 51.9  & 50.5  & 51.2  & 35.8  & 32.1  & 33.8  & 44.6  \\ \cline{2-15}
          
          & CSW & 54.8  & 44.4  & 49.1  & 42.6  & 44.1  & 43.3  & 50.8  & 49.4  & 50.1  & 44.0  & 32.3  & 37.3  & \underline{44.9}  \\ 
          & \multicolumn{1}{c||}{+tta} & 58.0  & 45.4  & 50.9  & 45.0  & 45.1  & \textbf{45.1}  & 54.4  & 50.8  & \textbf{52.6}  & 44.7  & 32.6  & \underline{37.7}  & \textbf{46.6} \\ \hline
          
          \multirow{5}{*}{\makecell[c]{m2m100\\ \citep{fan2021beyond}}} & CL & 11.1 & 4.6 & 6.5 & 20.7 & 12.4 & 15.5 & 14.5 & 8.3 & 10.5 & 44.0 & 30.6 & 36.1 & 17.1  \\ \cline{2-15}
          & CT & 52.7 & 47.0 & 49.7 & 31.9 & 33.8 & 32.8 & 47.3 & 51.1 & 49.1 & 31.2 & 31.8 & 31.5 & 40.8 \\ 
          & \multicolumn{1}{c||}{+tta} & 55.6 & 47.5 & \underline{51.2} & 39.1 & 36.9 & 38.0 & 49.7 & 54.4 & 51.9 & 34.0 & 32.7 & 33.3 & 43.6 \\ \cline{2-15}
          & CSW & 53.6 & 46.9 & 50.0 & 35.6 & 37.7 & 36.6 & 51.5 & 48.6 & 50.0 & 33.9 & 30.7 & 32.2 & 42.2 \\ 
          & \multicolumn{1}{c||}{+tta} & 53.8 & 47.2 & 50.3 & 43.5 & 43.2 & 43.3 & 53.4 & 51.3 & \underline{52.3} & 35.7 & 31.0 & 33.2 & 44.8 \\ \hline

      \end{tabular}
      \end{adjustbox}
      \caption{\label{tab:result}
      Main results on four datasets with different languages on cross-lingual ASTE task. wF1 scores are reported; the best results are in bold, and the second best are underlined. AVG represents the average wF1 score on all four datasets. ChatGPT-0 and ChatGPT-10 refer to zero-shot and 10-shot results of ChatGPT respectively. CL: cross lingual result; CT: complete translation, i.e., translate-train; CSW: code-switching. +tta refers to the results after combining our proposed test-time augmentation method. 
      }
    \end{table*}

    \begin{table*}[!ht]
      \centering
      \begin{adjustbox}{width=0.97\textwidth}
      \begin{tabular}{c|c||ccc|ccc|ccc||c}
      \hline
          \multicolumn{2}{c||}{} & \multicolumn{3}{c|}{Spanish} & \multicolumn{3}{c|}{Basque} & \multicolumn{3}{c||}{Catalan} & \multirow{2}{*}{\textbf{AVG}} \\
          \multicolumn{2}{c||}{} & NP-wP & NP-wR & NP-wF1 & NP-wP & NP-wR & NP-wF1 & NP-wP & NP-wR & NP-wF1 & \\ \hline
  
          \multirow{5}{*}{mT5-base} & CL & 48.31 & 31.10 & 37.84 & 36.72 & 28.95 & 32.38 & 46.73 & 37.86 & 41.83 & 37.35 \\ \cline{2-12}

          & dict\_csw (static) & 58.39 & 43.00 & 49.53 & 40.07 & 41.63 & 40.83 & 54.31 & 45.81 & 49.70 & 46.69 \\
          & dict\_csw (dynamic) & 59.18 & 44.10 & 50.54 & 40.54 & 37.39 & 38.90 & 55.56 & 43.91 & 49.05 & 46.16 \\ \cline{2-12}

          & CT & 56.31 & 45.41 & 50.28 & 46.36 & 44.39 & \textbf{45.36} & 55.43 & 52.57 & 53.96 & \underline{49.87}  \\ 
          & our CSW & 56.73 & 46.00 & 50.81 & 43.79 & 45.80 & \underline{44.77} & 55.47 & 53.90 & \textbf{54.67} & \textbf{50.08} \\ \hline
          
          \multirow{5}{*}{m2m100} & CL & 12.68 & 5.17 & 7.35 & 20.98 & 12.75 & 15.86 & 15.52 & 9.01 & 11.40 & 11.54  \\ \cline{2-12}

          & dict\_csw (static) & 36.86 & 19.12 & 25.18 & 23.05 & 14.46 & 17.77 & 33.13 & 18.67 & 23.88 & 22.28 \\
          & dict\_csw (dynamic) & 29.18 & 19.26 & 23.20 & 22.21 & 13.57 & 16.85 & 27.84 & 19.20 & 22.72 & 20.93 \\ \cline{2-12}

          & CT & 56.89 & 50.53 & \underline{53.52} & 35.61 & 36.98 & 36.28 & 51.95 & 56.14 & 53.96 & 47.92 \\ 
          & our CSW & 57.86 & 50.52 & \textbf{53.94} & 40.23 & 42.42 & 41.30 & 55.97 & 52.86 & \underline{54.37} & \underline{49.87} \\ \hline
  
      \end{tabular}
      \end{adjustbox}
      \caption{\label{tab:boundary_pred}
        Boundary prediction results with non-polar wF1 on three datasets with different languages on cross-lingual ASTE task. dict\_csw refers to the results of dictionary-based code-switching. Static and dynamic in the parentheses refer to different strategies for loading the bilingual dictionary. 
      }
    \end{table*}

    The main results are illustrated in Table \ref{tab:result}. Based on the results, we have the following observations:

    \subsubsection{Our TT-CSW framework boosts the performance of backbone models on cross-lingual ASTE.}
      As shown in Table \ref{tab:result}, the cross-lingual ASTE results of mT5-base and m2m100 are significantly improved after applying our proposed TT-CSW framework. Specifically, compare to the original cross-lingual results, the weighted-averaged F1 on Spanish, Basque, Catalan datasets are improved by 15.2\%, 13.9\% and 14.4\% respectively when using mT5-base as the backbone model. As for m2m100 backbone, the weighted-averaged F1 on Spanish, Basque, Catalan datasets are improved by 43.8\%, 27.8\% and 41.8\% respectively. This proves that the combination of training phase bilingual code-switching and testing phase alignment-based code-switching can significantly improve cross-lingual understanding of backbone models.
      
      As for the results on Norwegian dataset, we can observe some abnormal phenomena: all the models perform worse than the all-null baseline (i.e., outputs an empty list for all the test samples). We notice that 47\%
      of the test samples in Norwegian dataset do not contain any aspect or opinion terms. It is an unusual high rate of empty labels compared to the other three datasets, which are 11.7\%, 21.3\% and 16.1\% for Spanish, Basque and Catalan datasets respectively. We suspect that the labeling standard for Norwegian dataset is different from the other three datasets, which drops irrelevant aspects and opinions during the annotation process. 


    \subsubsection{Test time augmentation further improves performance.}
      As depicted in Table \ref{tab:result}, we can observe that the performance of complete translation (CT) and code-switching (CSW) are both improved after applying our proposed test-time augmentation method. For mT5-base backbone, the average wF1 on four datasets improved by 1.8\% and 1.6\%, as compared to the original CT and CSW results. As for m2m100 backbone, the improvements are 2.8\% and 2.6\% respectively. By combining training and testing phases of code-switching, we can achieve an improvement of 3.7\% and 4.0\% on average wF1 for mT5-base and m2m100 respectively. The bilingual multi-view of the input sentence introduced by our proposed test-time augmentation method can reduce the ambiguity of the input sentence, therefore further enhancing model performance. 
    
    \subsubsection{Our TT-CSW framework surpasses evaluation results of both ChatGPT and GPT-4.}
      We use the same zero-shot prompt as in \citet{gou-etal-2023-mvp} for cross-lingual ASTE task, which briefly describes the task and the definition of the output triplet first, and then provides the format for the output. As for the few-shot prompt, we randomly select 10 samples from the english training set and use them across all four datasets. The details of the prompts are listed in Appendix \ref{sec:prompts}. The results are listed in Table \ref{tab:result}. We can observe that our proposed TT-CSW framework outperforms ChatGPT on all four datasets. Specifically, the average wF1 score of our proposed TT-CSW framework with mT5-base as backbone model is 46.6\%, which is 15.7\% higher than ChatGPT-0 and 14.3\% higher than ChatGPT-10. This evidence indicates that even though ChatGPT can perform zero-shot cross-lingual transfer on ASTE tasks, its efficiency is still significantly less than that of smaller, fine-tuned models using our proposed TT-CSW framework. When it comes to GPT-4, except for the Spanish dataset, our TT-CSW framework outperforms its 10-shot performance. We surmise this could be because GPT-4 has extensive knowledge of the Spanish language, which is not the case for the other three languages with scarce annotation resources.

  \subsection{Boundary Prediction Analysis}
    To examine the effectiveness of our proposed boundary-aware code-switching method, we conduct a further analysis on the boundary prediction results on the Spanish, Basque and Spanish datasets. In specific, we use an evaluation metric called non-polar weighted-averaged F1 (NP-wF1). This metric is similar to wF1 as defined in section \ref{sec:eval_metrics}, except that we ignore the sentiment polarity part during the matching between predicted triplets and ground-truth triplets. In this way, we can focus on evaluating term boundaries. For dictionary-based code-switching, we use the bilingual dictionary released by \citet{qin2021cosda}, which is based on MUSE \citep{lample2018word}. We use two different strategies for loading the bilingual dictionary: static and dynamic. Static refers to the strategy that we construct the code-switched samples before the training phase, and the switched words are fixed during the training phase. Dynamic refers to the strategy that we reconstruct the code-switched samples at the start of each epoch during the training phase. We keep a ratio of 0.3 for the probability of switching each word based on the bilingual dictionary.

    The results are listed in Table \ref{tab:boundary_pred}. We can observe that for both mT5-base and m2m100, our proposed boundary-aware code-switching method outperforms the dictionary-based code-switching method and the complete translation method. For dictionary-based code-switching, static strategy performs better than dynamic strategy. Also, m2m100 struggles to predict terms accurately given the dictionary-based code-switched context. We suspect that this is due to the fact that bilingual dictionary contains some noisy translations, which may lead to incorrect term boundaries. Overall, the results prove the efficacy of improving term boundaries prediction with our proposed boundary-aware code-switching method.

  \subsubsection{Effect Analysis}
\begin{figure}[!t]
        \centering \includegraphics[width=0.99\linewidth]{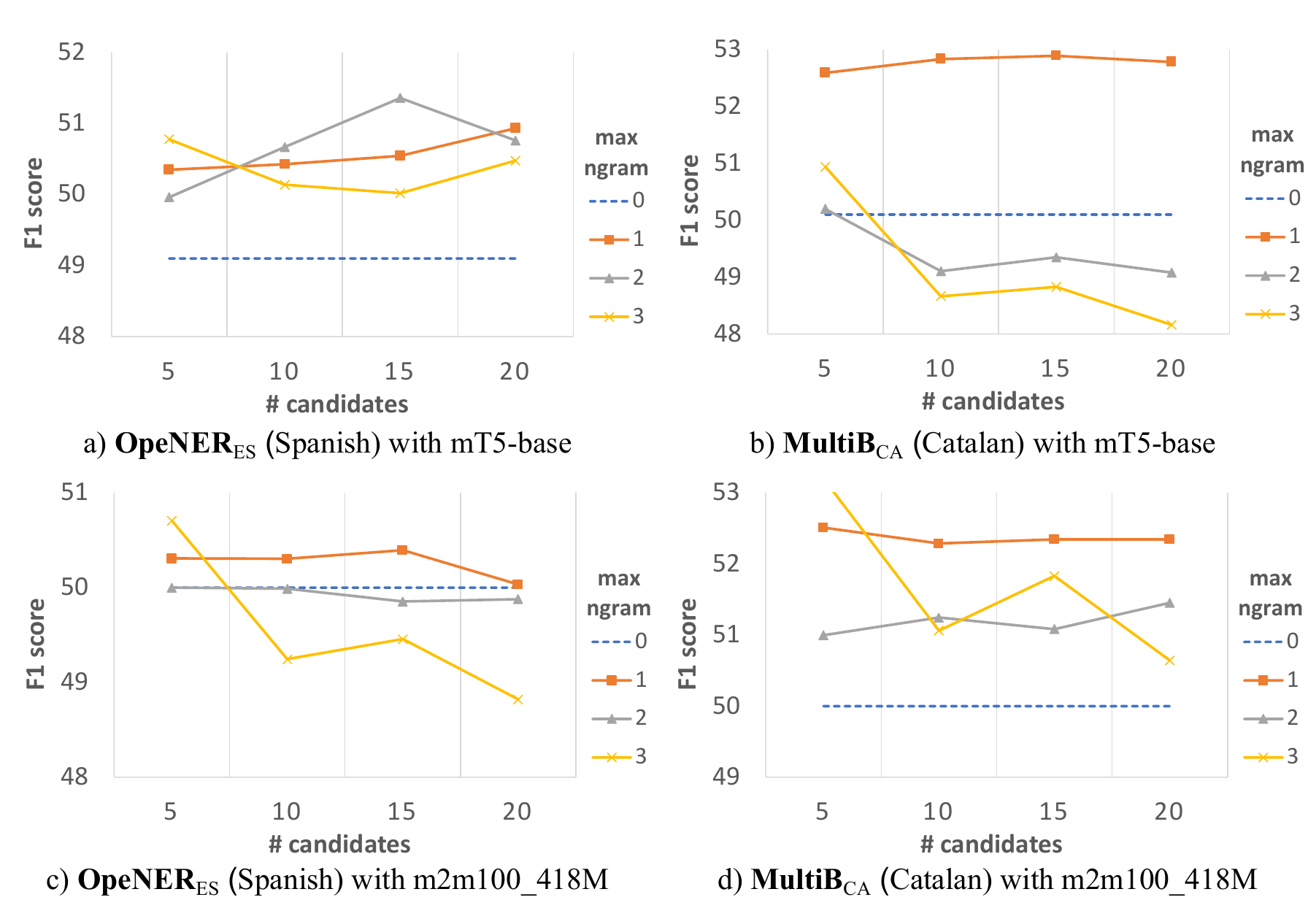}
        \caption{Effect of maximum n-gram and number of candidates for code-switching in test phase on Spanish and Catalan datasets. "\# candidates" refers to the number of augmented input sentence.}
        \label{fig:max_term_effect}
      \end{figure}

    We conduct an analysis on the effect of maximum n-gram and number of candidates for code-switching in test phase. The results are depicted in Figure \ref{fig:max_term_effect}. When the number of candidates is relatively small (i.e., 5), increasing maximum n-gram helps to improve performance. However, when the number of candidates is larger, the improvement is not stable and the performance even decreases. We suspect that this is because the number of candidates is already large enough to cover the possible code-switched context, and increasing maximum n-gram may introduce more noise. 

\section{Related Works}
 \subsection{Aspect Sentiment Triplet Extraction}
  Aspect Sentiment Triplet Extraction (ASTE) task was first proposed by \citet{peng2020knowing} as a subtask of aspect-based sentiment analysis (ABSA), and has drawn increasing attention in recent years with various kinds of methods. \citet{peng2020knowing}, \citet{xu2020position} and \citet{liang2023stage} proposed to use tagging-based approaches to solve this task. Graph-based encoding methods are also proposed for modeling the relationships between words \citep{barnes2021structured, chen2022enhanced}. There also exists some works that tried to formalize this task as a machine reading comprehension (MRC) task \citep{chen2021bidirectional, liu2022robustly, zhai2022mrc}. With the fashion of multi-task learning, generative methods are proposed to solve not only ASTE task, but also other ABSA subtasks together in a unified framework \citep{yan2021unified, zhang2021towards, gao2022lego, gou-etal-2023-mvp}.

 \subsection{Cross-lingual Transfer}
  Even though supervised methods have attained remarkable results on high-resource languages, it is still a challenge to attain comparable performance for languages with limited annotated resources. Cross-lingual transfer is one of the solution to this data scarcity issue on low-resource languages. It aims to solve the problem by leveraging the knowledge from high-resource languages \citep{schuster2019cross, lin2019choosing}. Existing methods on cross-lingual transfer can be roughly divided into two categories: data transfer and representation transfer. Data transfer methods usually rely on pesudo-labels on target language generated from machine translation tools \citep{fei2020cross, zhang2021cross} or knowledge-distillation methods \citep{liu2020cross, ge2023prokd}. Representation transfer methods try to align the representations of source and target languages in a shared space and exploit the language-independent features \citep{nooralahzadeh2020zero, huang2021improving, huang-etal-2023-pram}. Existing works on cross-lingual ABSA mainly focus on sentiment polarity part, which utilize translation-based methods \citep{barnes2016exploring, zhang2021cross} or teacher-student distillation \citep{lin2023cl}. However, few attempts have been made to apply cross-lingual transfer to ASTE task.

\section{Conclusion}
 In this study, we present a new code-switching framework for the cross-lingual aspect-based sentiment extraction (ASTE) task that can be easily incorporated with a variety of generative backbone models. It bridges the gap between the bilingual training phase and the monolingual test-time prediction. Our approach includes a boundary-aware code-switching method via the translation system, significantly improving the accurate determination of term boundaries. Additionally, we have designed a test-time augmentation alignment method that minimizes the ambiguity of the input sentence and further boosts model performance. Our proposed Test-time Code-Switching Framework (TT-CSW) has been thoroughly evaluated under four cross-lingual ASTE datasets with different languages, demonstrating its effectiveness. By integrating our method with several benchmark models, we attain an average improvement of 3.7\% on weighted F1-score. We also evaluate ChatGPT and GPT-4, two commonly used Large Language Model (LLM) developed by OpenAI. Furthermore, we prove that small generative models, when combined with our proposed TT-CSW framework, can exceed the performance of ChatGPT and GPT-4 by 14.2\% and 5.0\% respectively.

\section*{Limitations}
  Despite the promising results, our proposed TT-CSW framework still has some limitations for future work. Firstly, our proposed boundary-aware code-switching method relies on the translation system, which may introduce translation errors. Secondly, our proposed test-time augmentation method may introduce additional computational cost, which requires a trade-off between performance and efficiency for real-time applications. Lastly, we only evaluate our proposed framework on cross-lingual ASTE task, further experiments are needed to expand the scope of our proposed framework to other cross-lingual tasks.

\section*{Ethics Statement}
  Our experiments are conducted using publicly accessible datasets, ensuring no personal information is gathered. There's no utilization of sensitive or private data in our research processes. We maintain a strict policy against the use of any data that could potentially harm an individual, group, or the environment.


\bibliography{custom}

\newpage
\appendix

\section{Calculation of Evaluation Metrics}
  \label{sec:eval_metrics}
  We use the same type of symbols as defined in section \ref{sec:task_def}. The similarity score between a pair of triplets $T_1$ and $T_2$ can be calculated as shown in Equation \ref{func:eval1}, where $overlap(a, b)$ represents the number of overlapping words between a pair of string $a$ and $b$. If the sentiment polarity part is not correctly predicted, we consider it as an incorrect prediction and the similarity score is set to 0. We omit the respective terms in Equation \ref{func:eval1} when $OT_1$ or ${AT_1}$ is left empty. For the special case when terms in both triplets are empty, we regard it as an exact match. The weighted-averaged precision and recall score are calculated as shown from Equation \ref{func:eval2} to \ref{func:eval4} respectively. The $RT_i$ and $PT_i$ denote the ground truth triplet list and the predicted triplet list for sample $D_i$.

  \begin{equation}
    \label{func:eval1}
    \begin{aligned}
    sim(T_{1}, T_{2}) &= \frac{overlap(OT_1, OT_2)}{2len(OT_1)} \\
                  & + \frac{overlap(AT_1, AT_2)}{2len(AT_1)}
    \end{aligned}
  \end{equation}

  \begin{equation}
    \label{func:eval2}
    \begin{aligned}
    wP= \frac{\sum_{i=1}^N \sum_{T_j\in PT_i} \max_{T_k \in RT_i}(sim(T_j, T_k))}{TP+FP}
    \end{aligned}
  \end{equation}

  \begin{equation}
    \label{func:eval3}
    \begin{aligned}
    wR= \frac{\sum_{i=1}^N \sum_{T_j\in RT_i} \max_{T_k \in PT_i}(sim(T_j, T_k))}{TP+FN}
    \end{aligned}
  \end{equation}

  \begin{equation}
    \label{func:eval4}
    \begin{aligned}
    wF1=\frac{2wP\cdot wR}{wP+wR}
    \end{aligned}
  \end{equation}

\section{Prompts for ChatGPT and GPT-4}
  \label{sec:prompts}
  As shown in Listing \ref{lst:list1} and \ref{lst:list2}, we list the zero-shot and few-shot prompts for ChatGPT and GPT-4 in our experiments on the cross-lingual ASTE task. We use the same few-shot prompt across all the four datasets with different languages.

  \begin{figure*}
      \centering
\begin{lstlisting}[caption={Zero-shot Prompt for cross-lingual ASTE task.},basicstyle=\small, label=lst:list1]
According to the following sentiment elements definition: 

- The 'aspect term' refers to a specific feature, attribute, or aspect of a product or service that a user may express an opinion about.
- The 'opinion term' refers to the sentiment or attitude expressed by a user towards a particular aspect or feature of a product or service.
- The 'sentiment polarity' refers to the degree of positivity, negativity or neutrality expressed in the opinion towards a particular aspect or feature of a product or service, and the available polarities inlcudes: 'positive', 'negative' and 'neutral'.

Recognize all sentiment elements with their corresponding aspect terms, opinion terms and sentiment polarity in the following text with the format of [('aspect term', 'opinion term', 'sentiment polarity'), ...]: 
\end{lstlisting}
  \end{figure*}

  \begin{figure*}
      \centering
\begin{lstlisting}[caption={Few-shot Prompt (10 shots) for cross-lingual ASTE task.}, basicstyle=\small, label=lst:list2]
According to the following sentiment elements definition: 

- The 'aspect term' refers to a specific feature, attribute, or aspect of a product or service that a user may express an opinion about.
- The 'opinion term' refers to the sentiment or attitude expressed by a user towards a particular aspect or feature of a product or service.
- The 'sentiment polarity' refers to the degree of positivity, negativity or neutrality expressed in the opinion towards a particular aspect or feature of a product or service, and the available polarities inlcudes: 'positive', 'negative' and 'neutral'.

Recognize all sentiment elements with their corresponding aspect terms, opinion terms and sentiment polarity in the following text with the format of [('aspect term', 'opinion term', 'sentiment polarity'), ...]: 
Text: Although I wouldn 't say this was a cheap holiday , it didn 't break the bank either , so if you want a guaranteed tan go to Egypt , if not and its your main summer / yearly holiday , I wouldn 't recommended it ... I , d go somewhere else to avoid disappointment .
[('it', "wouldn 't recommended", 'negative'), ('somewhere else', 'd go', 'negative')]

Text: The Frankfurter Hof is surrounded by some of Europe 's most impressive skyscrapers .
[]

Text: Near hotel there are many bars , pubs , clubs .
[('clubs', 'Near hotel there are', 'positive'), ('many bars', 'Near hotel there are', 'positive'), ('pubs', 'Near hotel there are', 'positive')]

Text: Great central location
[('location', 'central', 'positive'), ('location', 'Great', 'positive')]

Text: Ofitsyanty while working well .
[('', 'working well', 'positive')]

Text: Never worth the money we had to pay for the room and the stay !!!
[('the money we had to pay', 'Never worth', 'negative')]

Text: You can pay exrra for these by paying for a ' gold all inclusive ' package
[]

Text: Very good parking possibilties .
[('parking possibilties', 'Very good', 'positive')]

Text: Nevertheless , because of its good location aside the liverpool One Shopping centre , with a lot of bars and restaurant , I continue going there when travelling to Liverpool .
[('location', 'good', 'positive'), ('location', 'continue going there', 'positive')]

Text: ( we had earplugs and used them !
[]
\end{lstlisting}
  \end{figure*}

\end{document}